# Sensitive Region-based Metamorphic Testing Framework using Explainable AI

Yuma Torikoshi
*The University of Electro-Communications*
Tokyo, Japan
yuma.torikoshi@uec.ac.jp

Yasuharu Nishi
*The University of Electro-Communications*
Tokyo, Japan
Yasuharu.Nishi@uec.ac.jp

Juichi Takahashi
AGEST, Inc.
Tokyo, Japan
juichi.takahashi@agest.co.jp

*Abstract*— Deep Learning (DL) is one of the most popular research topics in machine learning and DL-driven image recognition systems have developed rapidly. Recent research has employed metamorphic testing (MT) to detect misclassified images. Most of them discuss metamorphic relations (MR), with limited attention given to which regions should be transformed. We focus on the fact that there are sensitive regions where even small transformations can easily change the prediction results and propose an MT framework that efficiently tests for regions prone to misclassification by transforming these sensitive regions. Our evaluation demonstrated that the sensitive regions can be specified by Explainable AI (XAI) and our framework effectively detects faults.

## I. INTRODUCTION

Deep Learning (DL) driven image recognition system has been researched and implemented recently. Since the DL-driven image recognition system (IRS) is a black box system, metamorphic testing (MT) [1]–[3] is the most powerful test technique, which automatically generates various test cases. MT is a testing technique that automatically generates follow-up images for new test cases by transforming the seed image by specific relations.

In this paper, we propose another MT approach focusing on a sensitive region, where misclassification is more likely than others. Fig. 1 shows examples of the MT which adds a black dot. Example (a) is a seed image before adding a black dot, and it is classified as the gazelle by a DL-driven IRS trained by ImageNet [4]. Examples (b) and (c) are follow-up images after adding a black dot. Example (b) shows that the gazelle was misclassified as a lion when a 10 x 10 black dot was added over its face. On the other hand, example (c) shows that the image is correctly classified as a gazelle even when a 50 x 50 black dot is added to the background. They suggest that the position on the gazelle's face should be more sensitive than the position on the background to misclassify the seed image, i.e. to detect the bug, in the image recognition system.

Our MT approach aims at the sensitive region, where a higher failure detection rate (FDR) can be expected. FDR is defined as the percentage of the number of failures detected to the number of test cases executed. We also focus on the gradient to specify the sensitive region. Fig. 2 shows a heat map of the gradient by Grad-CAM [5], one of the Explainable AI (XAI) techniques. In Fig. 2(b) the red region indicates the highest gradient region while the blue regions indicate the lower gradient region. It suggests that the gradient should represent sensitivity.

According to the suggestions, our research questions are:

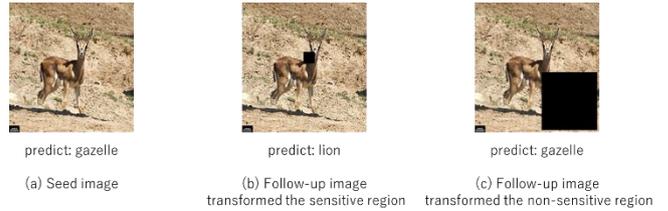

Fig. 1. Example of MT that add a black dot. A seed image classified as a gazelle (a). A follow-up image added a black dot inside the sensitive region (b). A follow-up image added a black dot outside the sensitive region(c).

RQ1. "Can Grad-CAM properly indicate sensitive regions?"

RQ2. "Is the FDR of our MT framework higher than the method that does not consider sensitivity?"

In this paper, section II shows the background and related works. Section III explains our MT approach. Section IV describes our MT method. In Section V we experiment to compare our MT framework and a traditional MT not considering sensitivity. Threats and validity are discussed in Section VI.

## II. BACKGROUND AND RELATED WORK

### A. Metamorphic Testing (MT)

Metamorphic Testing (MT) is a software testing technique that addresses the oracle problem [6], a common challenge in testing non-testable programs, where determining the expected output for a given input is difficult. MT utilizes metamorphic relations (MRs) to generate new test cases based on existing ones and to assess the software's reliability.

The procedure for MT is as follows. First, identify the System Under Testing (SUT)'s properties and establish MRs that describe the relationships between multiple inputs and their corresponding oracles. Second, generate an original set of test cases, execute them on the SUT, and obtain their seed image (house finch)     house finch     brambling     junco oracles. Third, apply the identified MRs to transform the original set of test cases into follow-up test cases, subsequently executing them on the SUT to acquire the corresponding oracles. Finally, verify whether the MRs hold for the original and follow-up test cases. If the MRs do not hold, a potential fault in the software is indicated.

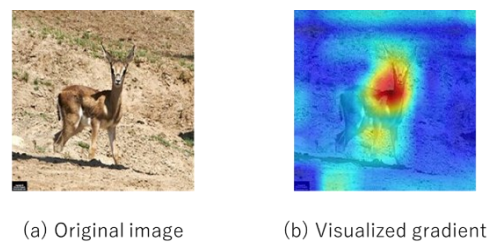

Fig. 2. Visualization of the gradient by Grad-CAM

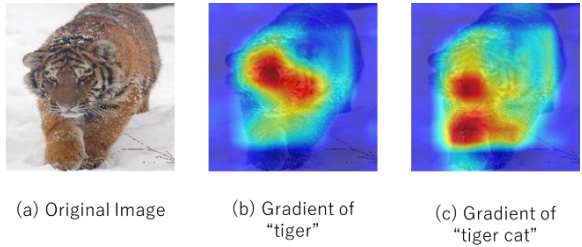

(a) Original Image     (b) Gradient of "tiger"     (c) Gradient of "tiger cat"

**Fig. 3. A visual explanation by Grad-CAM.**

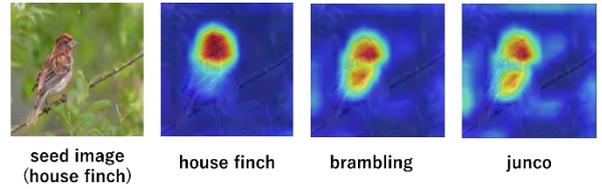

**Fig. 4. A specified gradient for each class to a seed image.**

In particular, in MT for IRS, an image of the original test case is transformed based on MRs to generate the image of the follow-up test case. There are two types of transformations based on MRs: transformations that transform the entire image (entire-based transformation) and transformations that transform a part of the region of the image (region-based transformation). Examples of entire-based transformations are affine transformations such as image rotation, scale, shear, and translation [7], as well as changing the weather or time of day in the road image [8]. Examples of region-based transformations are color conversions such as gray scaling, color inversion, brightness, etc. [9], [10], filtering transformations such as blurring and Gaussian noise [11], and data loss transformations such as adding black dots [12]. Among the region-based transformations, the one with precedent in previous research that transforms only a part of a region is the data loss transformation that adds black dots, and the region is selected randomly. We focus on the selection of the region to be transformed and propose the framework to efficiently generate follow-up test images that are more prone to misclassification than the random selection method.

*B. Explainable AI (XAI)*

When people make decisions, they explain their reasons, but the DL-System provides no reasons for prediction. It is dangerous to rely on the black box system without transparency. Therefore, AI that can explain why it makes the prediction, which is called Explainable AI (XAI), is a hot topic in research. In particular, XAI for convolutional neural networks (CNN) [13] has been widely studied, and XAI for CNNs can visually explain which pixel was used as the basis for the prediction.

Grad-CAM [5] is one of the XAI for CNN. It is a method that specifies the gradient which means the importance of the prediction as a heat map. It can provide us with the gradient of each class from an image by using the loss of probability of each class and the output of the last convolutional layer. Fig. 3 shows a visual explanation for the original image (Fig. 3.a). The red region on the heat map is a high-gradient region. The high gradient region which is mainly distributed on the face in the image (Fig. 3.b) means that the region had a strong influence on the prediction that the class is a tiger. On the other hand, the high gradient region which is mainly distributed on the legs in the image (Fig. 3.c) means that the region had a strong influence on the prediction that the class is tiger cat. Thus, Grad-CAM can provide the gradient heat map for each class for a single image.

Let the output value at the $(i,j)$ coordinates of the k-th feature map within the last convolutional layer be denoted as $A_k(i,j)$, and let the predicted probability for class c be $y^c$. The Grad-CAM's gradient $L_c(i,j)$ at the coordinate $(i,j)$ is computed using the following equation:

$$L_c(i,j) = ReLU\left(\sum_k \alpha_k^c A_k(i,j)\right) \quad (1)$$

Here, $\alpha_k^c$ represents the gradient indicating the extent to which the change at the coordinate $(i,j)$ affects the predicted probability for class c. When $Z = \sum_i \sum_j 1$, $\alpha_k^c$ is calculated using the following equation:

$$\alpha_k^c = \frac{1}{Z}\sum_{i,j} \frac{\partial y^c}{\partial A_k(i,j)} \quad (2)$$

In this paper, $L_c$ is referred to as the gradient for class c.

Although there are several other XAI methods for CNN, Fan et al. showed that Grad-CAM is one of the best XAI methods. [14]

III. APPROACH

Our approach aims to efficiently test regions with a high likelihood of misrecognition by applying existing natural metamorphic transformations, thereby enabling reliable testing. The transformation of images is determined by which region is changed and how the region is changed. Generally, MT transforms the region naturally, regardless of which regions are altered, to test for reliability. For instance, tests are conducted by adding black dots to verify whether misclassification occurs in natural situations, such as when leaves or dust appear on the image. Additionally, tests involving brightness changes or gaussian noise assess whether misclassification occurs in natural situations with environmental light or camera noise. Thus, MT typically involves the addition of natural situational changes to test reliability.

Our method adopts an approach that efficiently tests regions prone to misclassification by determining which regions to change using Explainable Artificial Intelligence (XAI). While it is technically possible to determine how the region is changed so that it is more likely to be misclassified, such transformations are called adversarial attacks and test security rather than reliability. Consequently, our study exclusively focuses on an approach that utilizes XAI to determine which regions to transform.

Grad-CAM is a technology to calculate the importance of prediction by gradients. It can calculate which regions are highly sensitive regions that had a significant impact on the predictions. For example, the red region in Fig. 2 (b) which shows the highest gradient region is the most sensitive. Our approach is "more gradient, more misclassification".

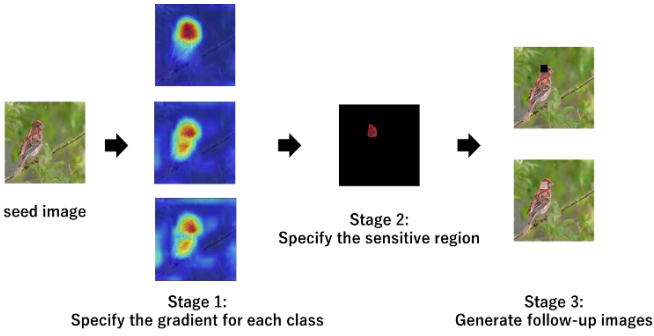

Fig. 5. Overview of our framework.

Grad-CAM should hence be applied to MT. If our method transforms a high gradient region specified by Grad-CAM, misclassification is likely to happen. We call it Sensitive Region-based MT, (Sensitivity-MT).

IV. METHODS

### A. Overview

The overall design of our framework is shown in Fig. 5. Our framework consists of three stages. First, in Stage 1, gradients for each class are identified as a heat map. Next, in Stage 2, we specify sensitive regions for transformation based on the gradient heat maps specified in Stage 1. There are three methods of specifying the sensitive region. (Three methods are discussed in more detail below (IV.C)). Finally, in Stage 3, rectangle regions for transformation are decided based on the sensitive region, and the rectangle regions are transformed based on the MR.

### B. Stage 1: Specifying the gradient for each class

In Stage 1, we simply apply Grad-CAM to the seed image and specify the gradient for each class. Fig. 4 shows an example of identified gradients for each class from a seed image of a house finch class. The gradients are obtained as a heat map. This means that Lc in equation (1) is obtained for all classes c.

### C. Stage 2: Specifying the sensitive region

Since gradients were identified for each class in stage 1, there are as many gradient heat maps as there are classes. In Stage 2, one sensitive region is specified based on the gradients. It is hoped that transformation within the sensitive region would make it prone to misclassification. When specifying the sensitive region from gradients, it is not clear which class's gradients are more important and how should we specify the sensitive region. Therefore, we propose three Sensitivity-MTs that use different specifying sensitive region methods. The three methods are named the max selection method, the avg selection method, and the best selection method. The three methods generate heat maps of the sensitive regions.

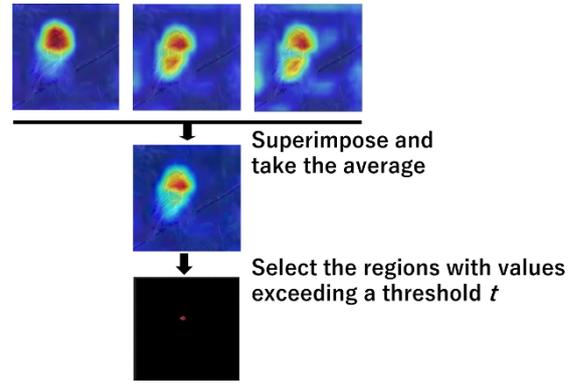

Fig. 7. A workflow for selecting the region for transformation using the average selection method.

*1) The max selection method (Max Selection)*

The max selection method superimposes heat maps of gradients and takes the maximum value, as shown in Fig. 6. Each pixel on the heat map has a value between 0 and 1 representing the gradient, with blue color as it approaches 0 and red as it approaches 1. A heat map of the sensitive region is generated by taking the maximum gradient for each pixel. Pixels on the heat map with importance values less than a gradient threshold are removed from the sensitive region. The larger the threshold, the smaller the sensitive region.

When the threshold is $t$ and the set of all classes is $C$, the sensitive region ($R_{sensitive}$) is represented by the set of sensitive coordinates as follows.

$$R_{sensitive} = \left\{(i,j) \mid t \leq \max_{c \in C}\left(L_c(i,j)\right)\right\} \quad (3)$$

This approach is based on the idea that the region where gradients of all classes are high should be sensitive regions.

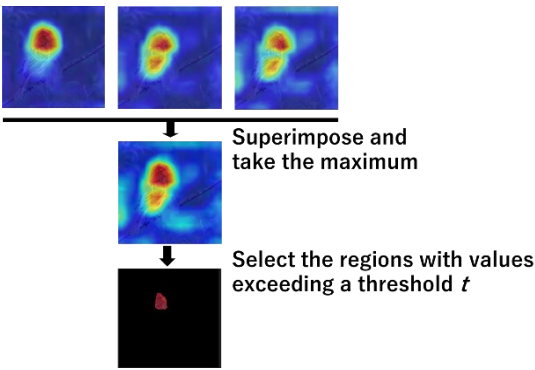

Fig. 6. A workflow for selecting the region for transformation using the max selection method.

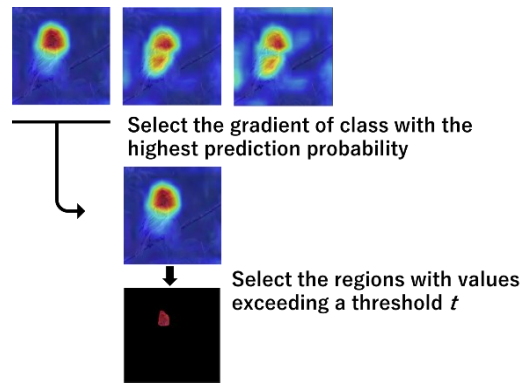

Fig. 8. A workflow for selecting the region for transformation using the best selection method.

### 2) The Average selection method (Avg Selection)

The average selection method superimposes heat maps of gradients and takes the average value, as shown in Fig. 7. The Average selection method differs from the Max selection method only in that it takes the average value instead of the maximum value.

The sensitive region ($R_{sensitive}$) is represented by the set of sensitive coordinates as follows.

$$R_{sensitive} = \left\{(i,j) \mid t \leq \underset{c \in C}{\text{mean}} \left(L_c(i,j)\right)\right\} \quad (4)$$

This approach is based on the idea that the regions with high gradients common to many classes should be the sensitive regions.

### 3) The best selection method (Best Selection)

The best selection method selects a heat map of the class with the highest prediction probability as shown in Fig. 8. Pixels on the heat map with importance values less than the gradient threshold are removed from the sensitive region.

When the class with the highest prediction probability is $c_{best}$, the sensitive region ($R_{sensitive}$) is represented by the set of sensitive coordinates as follows.

$$R_{sensitive} = \left\{(i,j) \mid t \leq L_{c_{best}}(i,j)\right\} \quad (5)$$

This approach is based on the idea that the region where the gradient of the class with the highest prediction probability is high should be the sensitive region because decreasing the prediction probability of the class with the highest prediction probability can lead to misclassification.

### D. Stage 3: To decide on regions for transformation and transform the regions.

In Stage 3, transformations are based on MR. The candidate rectangle regions for transformation are determined as shown in the blue box in Fig. 9. This is based on the $R_{sensitive}$ generated in Stage 2.

The rectangle region has arbitrary width and height based on each pixel in the red area of Fig. 9. If all pixels in the $R_{sensitive}$ are used as a base pixel, the number of rectangle regions would be too large, so a stride is set to reduce the number of base pixels as shown in Fig. 9.

For transformations based on MR, the method proposed in previous studies that can transform any one region can be applied as is. For example, the methods of adding black dots and transforming color tones can be used.

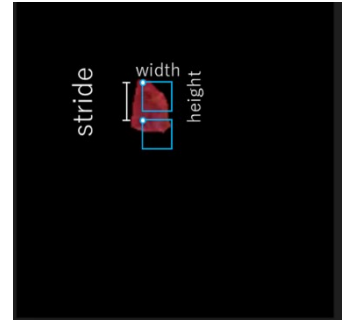

Fig. 9. Decide the rectangle regions to be transformed from the sensitive region.

## V. EXPERIMENTS

### A. Preparation

We conducted experiments to evaluate two RQs. Before the evaluation, we prepared an image recognition AI as SUT. It is a pre-trained VGG16 [15] model in Keras 2.10.0, which classifies 999 classes of images in ImageNet [4] (class n04399382" is missing due to ImageNet). We prepared 999 seed images which are correctly classified into different classes by the SUT. The experiment worked on a PC with an Intel Core i7 10400F CPU, RTX 3060 Ti GPU, and 16 GB RAM.

Misclassification by MT is generally affected by the number of regions, position of regions, size of regions, and type of metamorphic transformation. The purpose of our experiment is to validate how sensitivity affects misclassification by comparing the positions of transformed regions between a baseline MT method and our sensitivity-based MT framework. We selected a baseline MT method, the Random Selection method from DeepXplore[12], which randomly selects the positions of regions to be transformed. Although DeepXplore originally selects multiple regions, the number of regions for the baseline and our framework should be limited to the same number of regions, i.e. limited to one single region in this experiment. We specified that the size of regions is 10 x 10 pixels, and picked color inversion, hole, brightness, blurring, and Gaussian noise as types of metamorphic transformation. Fig. 10 shows examples of the transformation. The thresholds in our three Sensitivity-MTs were 0.9 and the strides were 5 pixels.

### B. Correlation analysis between sensitivity and FDR (RQ1)

To evaluate RQ1, "Can Grad-CAM properly indicate sensitive regions?", we should analyze the correlation between the sensitivity and the FDR in our three Sensitivity-MTs without the baseline. We analyzed the correlation between the gradient, i.e. approximated sensitivity in various positions of regions. Gradients were calculated in 2,497,500 regions, i.e. randomly selected 500 regions in 4,995 follow-up images generated by five types of metamorphic transformation for 999 seed images. The gradients of a region were represented by the gradients of a pixel at the center of the region.

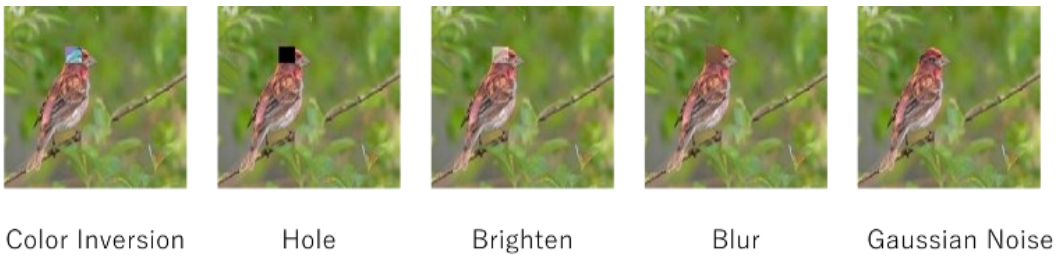

Fig. 10. Five types of region-based metamorphic transformation.

Table. 1. Comparison of FDRs for the baseline and three Sensitivity-MTs

|  |  | # of positive | # of negative | FDR | FDR/ Baseline's FDR |
|---|---|---|---|---|---|
| Baseline | Random Selection | 2360695 | 136805 | 5.48% | - |
| Ours | Max Selection | 730793 | 75787 | **9.40%** | 1.89 |
|  | Avg Selection | 39369 | 3141 | **7.39%** | 1.72 |
|  | Best Selection | 76456 | 8849 | **10.37%** | 1.35 |

### C. Comparison of the baseline FDR with the FDRs of our three Sensitivity-MTs (RQ2)

To evaluate RQ2: "Is the FDR of our MT framework higher than the method that does not consider sensitivity?", we compared the FDR of the baseline method and the FDR of the three Sensitivity-MTs in the follow-up test images generated from 999 seed images.

## VI. RESULTS

### A. RQ1: Can Grad-CAM properly indicate sensitive regions?

The answer to *RQ1: Can Grad-CAM properly indicate sensitive regions?* is positive in this experiment.

Fig. 11 shows the correlation between the gradients in the regions and the FDR by each Sensitivity-MT. All the correlation coefficients are significantly high, that is, 0.982 for Max Selection, 0.962 for Avg Selection, and 0.904 for Best Selection. All the FDRs are monotonically increased except for the short range. The result suggests that the larger the gradient of a region, the more likely to be misclassification when the region is transformed. In other words, a high-gradient region is a sensitive region.

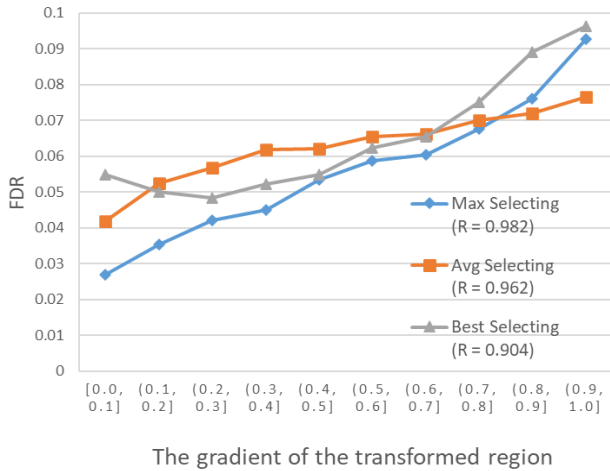

**Fig. 11. A relationship between the sensitivity of the transformed region and the FDR with three Sensitivity-MTs**

### B. RQ2: Is the FDR in our MT framework higher than the method that does not consider sensitivity?

The answer to *RQ2: Is the FDR in our MT framework higher than the method that does not consider sensitivity?* is positive in this experiment.

Table. 1 shows the results of the comparison between the FDR of the baseline and that of our three Sensitivity-MTs. The # of positives indicates the number of images correctly classified, and The # of negatives indicates the number of images misclassified. FDR is calculated as the percentage of the number of negatives out of the total number of positives and negatives. The FDR/Baseline's FDR is defined as the ratio of the baseline FDR. All Sensitivity-MTs displayed a higher FDR than the baseline. The Best Selection Sensitivity-MT demonstrated the highest FDR, at 10.37%, which is 1.89 times greater than the baseline FDR.

## VII. THREATS TO VALIDITY

This framework effectively generates images that are misclassified by utilizing Grad-CAM. Although this is efficient for the FDR, it is not necessarily efficient in terms of time spent; if Grad-CAM requires a longer duration to identify gradients, the random selection method may prove more efficient in the number of negatives discovered per hour. The time needed to specify gradients for all classes using Grad-CAM on a single seed image increases proportionally to the number of classes. The Best Selection method, which designates the crucial region for one class, required approximately 0.2 milliseconds for a single seed image, while the Max and Avg Selection methods, which determine gradients for all classes (1000 classes), necessitated approximately 17 seconds. Considering time efficiency, the Best Selection method emerges as the most efficient approach. Consequently, time efficiency constitutes a vital aspect and an important challenge to address.

## VIII. CONCLUSION AND FUTURE WORKS

In this paper, we proposed a framework that tests and detects misidentified images more efficiently using MT that transforms a region of the image. This framework can be combined with region-based MT from prior research, enabling focused testing on sensitive regions prone to misclassification. This is the first approach to focus on the sensitivity of the region on MT by using XAI, and we believe that this is the first step in research into the use of sensitivity for MT using XAI.

In addition to image recognition AI, we plan to use sensitivity for video recognition AI and other applications in the future.

REFERENCES

[1] T. Y. Chen, "Metamorphic testing: a simple method for alleviating the test oracle problem," in *Proceedings of the 10th International Workshop on Automation of Software Test*, Florence, Italy, May 2015, pp. 53–54.

[2] S. Segura, G. Fraser, A. B. Sanchez, and A. Ruiz-Cortés, "A Survey on Metamorphic Testing," *IEEE Trans. Softw. Eng.*, vol. 42, no. 9, pp. 805–824, Sep. 2016,

[3] T. Y. Chen *et al.*, "Metamorphic Testing: A Review of Challenges and Opportunities," *ACM Comput. Surv.*, vol. 51, no. 1, p. 4:1-4:27, Jan. 2018,

[4] O. Russakovsky *et al.*, "ImageNet Large Scale Visual Recognition Challenge," *Int. J. Comput. Vis.*, vol. 115, no. 3, pp. 211–252, Dec. 2015,

[5] R. R. Selvaraju, M. Cogswell, A. Das, R. Vedantam, D. Parikh, and D. Batra, "Grad-CAM: Visual Explanations from Deep Networks via Gradient-Based Localization," in *2017 IEEE International Conference on Computer Vision (ICCV)*, Oct. 2017, pp. 618–626.

[6] H. Liu, F.-C. Kuo, D. Towey, and T. Y. Chen, "How Effectively Does Metamorphic Testing Alleviate the Oracle Problem?," *IEEE Trans. Softw. Eng.*, vol. 40, no. 1, pp. 4–22, Jan. 2014,

[7] Y. Tian, K. Pei, S. Jana, and B. Ray, "DeepTest: Automated Testing of Deep-Neural-Network-Driven Autonomous Cars," in *2018 IEEE/ACM 40th International Conference on Software Engineering (ICSE)*, May 2018, pp. 303–314.

[8] M. Zhang, Y. Zhang, L. Zhang, C. Liu, and S. Khurshid, "DeepRoad: GAN-Based Metamorphic Testing and Input Validation Framework for Autonomous Driving Systems," in *2018 33rd IEEE/ACM International Conference on Automated Software Engineering (ASE)*, Sep. 2018, pp. 132–142.

[9] H. Spieker and A. Gotlieb, "Adaptive metamorphic testing with contextual bandits," *J. Syst. Softw.*, vol. 165, p. 110574, Jul. 2020,

[10] M. Pu, M. Y. Kuan, N. T. Lim, C. Y. Chong, and M. K. Lim, "Fairness Evaluation in Deepfake Detection Models using Metamorphic Testing," in *2022 IEEE/ACM 7th International Workshop on Metamorphic Testing (MET)*, May 2022, pp. 7–14.

[11] A. Wildandyawan and Y. Nishi, "Object-based Metamorphic Testing through Image Structuring." arXiv, Feb. 12, 2020.

[12] K. Pei, Y. Cao, J. Yang, and S. Jana, "DeepXplore: automated white box testing of deep learning systems," *Commun. ACM*, vol. 62, no. 11, pp. 137–145, Oct. 2019,

[13] Z. Li, F. Liu, W. Yang, S. Peng, and J. Zhou, "A Survey of Convolutional Neural Networks: Analysis, Applications, and Prospects," *IEEE Trans. Neural Netw. Learn. Syst.*, vol. 33, no. 12, pp. 6999–7019, Feb. 2022,

[14] M. Fan, J. Wei, W. Jin, Z. Xu, W. Wei, and T. Liu, "One step further: evaluating interpreters using metamorphic testing," in *Proceedings of the 31st ACM SIGSOFT International Symposium on Software Testing and Analysis*, New York, NY, USA, Jul. 2022, pp. 327–339.

[15] K. Simonyan and A. Zisserman, "Very Deep Convolutional Networks for Large-Scale Image Recognition." arXiv, Apr. 10, 2015.